\documentclass[10pt,twocolumn,letterpaper]{article}

\usepackage[pagenumbers]{cvpr} 

\usepackage{graphicx}
\usepackage{amsmath}
\usepackage{amssymb}
\usepackage{booktabs}
\usepackage{float}
\usepackage[capitalize]{cleveref}
\crefname{section}{Sec.}{Secs.}
\Crefname{section}{Section}{Sections}
\Crefname{table}{Table}{Tables}
\crefname{table}{Tab.}{Tabs.}

\def\cvprPaperID{4148} 
\def\confName{CVPR}
\def\confYear{2022}

\begin{document}

\title{
``The Pedestrian next to the Lamppost''\\ Adaptive Object Graphs for Better Instantaneous Mapping}
\author{Avishkar Saha$^{1}$, Oscar Mendez$^{1}$, Chris Russell$^{2}$, Richard Bowden$^{1}$\\
$^{1}$Centre for Vision Speech and Signal Processing, University of Surrey, Guildford, UK \\ $^{2}$\textit{Amazon}, Tubingen, Germany \\
{\tt\small \{a.saha, o.mendez, r.bowden\}@surrey.ac.uk, \tt\small cmruss@amazon.com}
}
\maketitle
\vspace{-0.3cm}
\begin{abstract}

Estimating a semantically segmented bird's-eye-view (BEV) map from a single image has become a popular technique for autonomous control and navigation. However, they show an increase in localization error with distance from the camera. While such an increase in error is entirely expected -- localization is harder at distance -- much of the drop in performance can be attributed to the cues used by current texture-based models, in particular, they make heavy use of object-ground intersections (such as shadows) \cite{dijk2019neural}, which become increasingly sparse and uncertain for distant objects.
In this work, we address these shortcomings in BEV-mapping by learning the spatial relationship between objects in a scene. We propose a graph neural network which predicts BEV objects from a monocular image by spatially reasoning about an object within the context of other objects. Our approach sets a new state-of-the-art in BEV estimation from monocular images across three large-scale datasets, including a 50\% relative improvement for objects on nuScenes. 

\end{abstract}
\vspace{-0.6cm}
\section{Introduction}
\label{sec:intro}
The ability to generate top-down birds-eye-view maps from images is an important problem in autonomous driving. Overhead maps provide compact representations of a scene's spatial configuration and other agents, making it an ideal representation for downstream tasks such as navigation and planning. Given their utility, the BEV estimation problem of inferring semantic BEV maps from images, has drawn increasing attention in recent years --- mapping `things' such as traffic cones and pedestrians, and `stuff' such as road lanes and pavements.

Current BEV estimation techniques \cite{Roddick_2020_CVPR, saha2021enabling, saha2021translating, philion2020lift} have made impressive strides towards high-accuracy semantic maps for both `things' and `stuff' from a single image. These texture-based models are elegant in their simplicity, requiring only minimal losses on their predicted BEV maps. Although these models work well for large amorphous textured classes that dominate the scene, such as road and pavement (a.k.a. stuff\cite{caesar2018coco}), they suffer from low recall and large localization error for smaller and potentially dynamic objects at greater distances (a.k.a. things). 

In contrast, the field of monocular 3D detection displays far greater object localization accuracy by taking an object-based approach. A simple solution to increase recall and localization accuracy in BEV estimation is to apply an off-the-shelf monocular 3D detector to generate BEV object bounding boxes. Surprisingly, this increases object intersection-over-union (IoU) accuracy on the BEV estimation task by a relative 20\%. This raises the question: why not use the best of both methods? That is, reason about objects in object space, and use them to improve the estimates of background `stuff'. 

We propose a novel BEV estimation method that leverages object graphs to reason about scene layout. These graphs provide a rich source of additional information to improve  object localization, as they generate context by propagating between objects. Our model predicts BEV objects from a monocular image by spatially reasoning about an object given the long-range context of other objects in the scene. The contributions of our work are as follows:
\begin{enumerate}
    \item We propose a novel application of graph convolution networks for spatial reasoning to localize BEV objects from a monocular image. 
    \vspace{-0.3cm}
    \item We demonstrate the importance of learning both node and edge embeddings, their mutual enhancement and edge supervision for the problem of object localization.
    \vspace{-0.3cm}
    \item We introduce positional-equivariance into our graph propagation method, leading to state-of-the-art results across three large-scale datasets, including a 50\% relative improvement in BEV estimation of `things' or objects.
\end{enumerate}

\section{Related Work}
\label{sec:related_work}
\textbf{BEV Estimation:} 
Initial work in road layout estimation \cite{sengupta2012automatic} took a two-staged approach of semantic segmentation in the image followed by a homography-based mapping to the ground plane. Others \cite{wang2019parametric, liu2020understanding, schulter2018learning} similarly exploited depth and semantic segmentation maps to lift scene and object entities into BEV. As these methods required dense annotation maps as additional input, recent work implicitly reasons about depth and semantics within the network. Some learn image-to-BEV transformations implicitly \cite{lu2019monocular, mani2020monolayout, pan2020cross}, while more recent approaches improved results by conditioning the transformation on the camera's geometry \cite{Roddick_2020_CVPR, saha2021translating, saha2021enabling, philion2020lift, dwivedi2021bird}. These methods can be broadly categorized by their transformation as either fixed or adaptive. Fixed approaches \cite{Roddick_2020_CVPR} use a fully-connected layer to vertically condense image features into a bottleneck and then expand into BEV with another fully-connected (FC) layer. One limitation of the approach is that the weights of both FC layers are fixed, increasing the layer's tendency to ignore small objects and resulting in a low recall for these categories. Adaptive approaches \cite{saha2021translating, philion2020lift} on the other hand, use an attention mechanism that generates context by querying a spatial location. The primary challenge of this approach is the size of the image search space: learning to select the appropriate image features for distant objects whose relevant context is sparse and uncertain is challenging. We overcome the issues with large search spaces by specifically communicating between objects as nodes, aided by scene context sampled along edges.




\color{black}
\textbf{Monocular Object Detection:} In contrast to BEV Estimation, monocular object detection methods are generally object-based, with losses applied per object. Much of this work is centered around constraining the search space --- both in the search for objects in the image and their 3D pose regression. A common and successful approach is to constrain the image search space via 2D object detection followed by 3D pose regression \cite{wang2019monocular, mousavian20173d, kehl2017ssd, simonelli2019disentangling, poirson2016fast}. Some approaches use geometric priors to constrain the projected 3D bounding box to fit within the 2D box \cite{mousavian20173d, gustafsson2018automotive}, while others leveraged the relationship between 2D box height and the estimated object height to create initial centroid proposals \cite{ku2019monocular}. Mono3D \cite{chen2016monocular} on the other hand generated 3D proposals on the ground-plane which are then scored by projecting back into the image. The primary drawback of all these methods is that each object proposal is generated independently. Some approaches have tried to reason globally across the scene. OFTNet \cite{roddick2018orthographic} built on Mono3D to reason globally in BEV by projecting a 3D voxel grid onto the image to collect image-features. However, this creates a bottleneck similar to the fixed BEV-estimators as the features assigned to a voxel are independent of depth. MonoPair \cite{chen2020monopair} constrained the positions of objects by optimizing their pairwise spatial relationships, however this was a post-optimization step outside the network using only the predicted 3D bounding boxes. While graphs have been used for semantic reasoning \cite{yang2018graph, dhamo2020semantic, liu2018structure}, unlike all previous work, we: (1) use graphs to reason about scene layout and (2) generate context by propagating between objects.

\textbf{Graph neural networks:} Graph neural networks (GNNs) have emerged as a powerful neural architecture to learn from graph-structured data, exhibiting promising results in social networks \cite{monti2019fake}, 
drug design \cite{stokes2020deep} and more. GNNs build node representations by aggregating local information from their neighborhood. Drawing on the successes of convolutional neural networks (CNNs), several works have generalized convolution operations to the graph domain. These graph convolution networks (GCNs) generally fall into two major categories: spectral and spatial. Spectral approaches \cite{defferrard2016convolutional, DBLP:conf/iclr/KipfW17} perform convolutions in the Fourier domain, while spatial approaches \cite{wu2018moleculenet, hamilton2017inductive} perform them in the node (or vertex) domain. Importantly, our graphs have Euclidean interpretations, thus our graph convolutions need to be spatial instead of spectral. Recently, Velickovic et al. \cite{velivckovic2018graph} developed an attention-based neighborhood aggregation mechanism operating in the node domain --- the Graph Attention Network (GAT). Here, every node updates its representation by attending to its neighbors conditioned on itself, resulting in GATs becoming the state-of-the-art neural architecture for graph learning \cite{bronstein2021geometric}.

Apart from a few notable exceptions \cite{coembedding, pmlr-v129-yang20a}, most graph learning methods ignore the representational capabilities of edges and focus only on node embeddings. CensNet \cite{coembedding} addressed the utility of edge features by simultaneously embedding both nodes and edges into a latent feature space using spectral graph convolutions. NENN \cite{pmlr-v129-yang20a} approached this co-embedding in the spatial domain. We similarly learn both node and edge embeddings along with their mutual enhancement. However, unlike all previous approaches, our graph propagation method is made position equivariant to account for the importance of Euclidean structure in localization.

\color{black}





\begin{figure*}[t]
\centerline{\includegraphics[page=2,trim=150 280 120 280,clip,width=1.\linewidth]{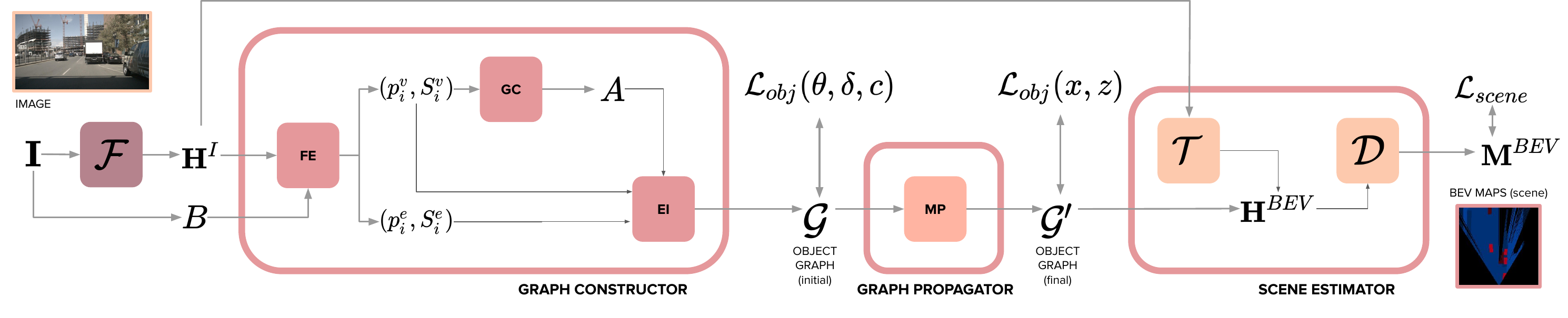}}
\vspace{-0.3cm}
\caption{Model Architecture. Our \textbf{Graph Constructor} takes image features $\mathbf{H}^I$ and candidate regions $B$ as input and generates a graph $\mathcal{G = (V, E)}$ of the scene's $N$ objects. The \textbf{Graph Propagator} passes messages across the nodes and edges of the graph in a position-equivariant manner, co-embedding both nodes and edges to a latent feature space for object localisation, resulting in a graph $\mathcal{G' = (V', E')}$ with updated embeddings. The \textbf{Scene Estimator} takes the image features $\mathbf{H}^I$ and transforms them to BEV where they are combined with object embeddings $\mathcal{V'}$ to generate BEV maps $\mathbf{M}^{BEV}$.}
\label{fig:model_arch}
\vspace{-0.5cm}
\end{figure*}

\section{Our approach}
\label{sec:method}

Given an image captured while driving, we wish to infer a semantically segmented BEV map of its scene. 
As shown in Fig.~\ref{fig:model_arch}, we approach BEV estimation as a two stage process of object localization followed by subsequent complete BEV estimation, including the localization of amorphous 'stuff' such as road or sidewalk that can not be treated as objects. To improve object localization  beyond the standard accuracies offered by 3D object detection, we  generate context by propagating information between objects to infer their spatial layout. 
In this section, we first motivate the use of graphs for object localization (Sec.~\ref{sec:graph_motivation}). We then discuss the main components of our approach shown in Fig.~\ref{fig:model_arch}: first, we structure scene content in the form of a graph across the image and/or BEV-plane (Sec. \ref{sec:graph_construction}). Next, we localize objects in BEV by propagating structural and positional embeddings across the graph (Sec. \ref{sec:map_estimation}). Finally, we use its learned embeddings to generate complete maps including `stuff' the amorphous regions such as road  (Sec. \ref{sec:map_estimation}).

\subsection{Design Motivations} \label{sec:graph_motivation}
\textbf{Why graph representations?} A natural question is what makes object graphs more suitable for localization than current BEV estimation methods?
The answer is twofold: (1) graphs encode explicit geometric relationships between objects that BEV networks have to model implicitly and (2) they allow nonlocal communication between entities, which a convolutional BEV network requires many downsampling operations to do. In this section, we build on this to motivate our need for (1) a graph which jointly learns node and edge embeddings; 
and (2) supervising edge embeddings as a way of placing geometric constraints.

\textbf{Why learn node embeddings?} 
The primary challenge in object localization is depth estimation.  
To resolve an object's depth, monocular scene understanding methods typically rely on visual cues from the object's intersection with the ground \cite{dijk2019neural}. 
While artefacts such as shadows  provide strong priors, they become increasingly sparse and uncertain for distant objects.
When such image features become unreliable, it remains possible to localize an object by comparing its appearance, geometry and position to others in the scene. This can be done by representing each object as a node in a graph and  message passing between them.

\textbf{Why learn edge embeddings?} While propagating between nodes provides a baseline mechanism for  localization, regressing a node's location within the graph still requires traversing a large search space. To restrict this space, we can place geometric constraints on object localization by predicting the midpoint of each edge --- however, this requires learned edge embeddings. 


\begin{figure*}[t]
\centerline{\includegraphics[page=1,trim=40 90 35 80,clip,width=0.8\linewidth]{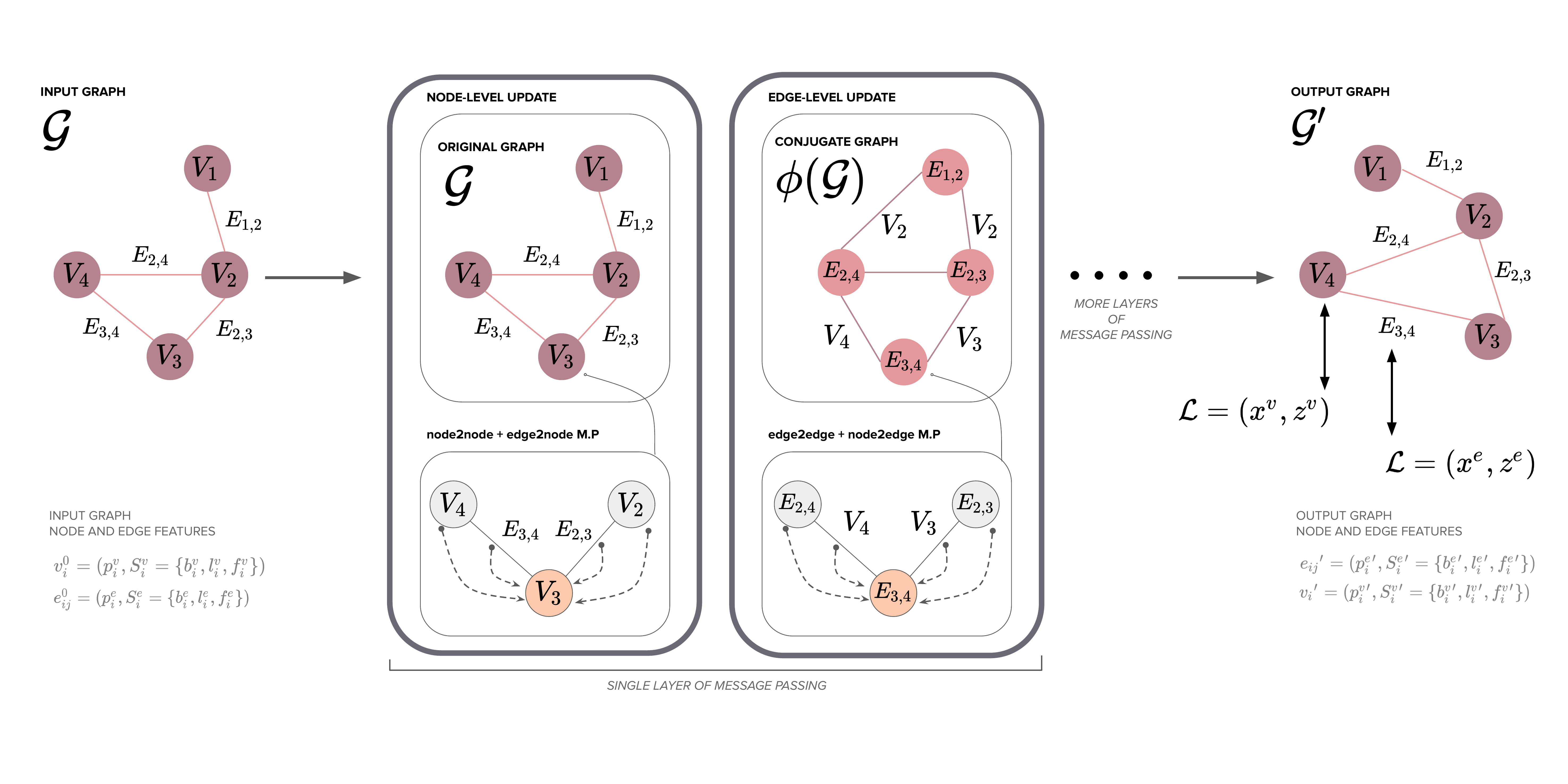}}
\vspace{-0.3cm}
\caption{Our graph propagation method. A single layer of message passing consists of node-to-node and edge-to-node updates, followed by edge-to-edge and node-to-edge updates. The node and edge embeddings of the output graph are then supervised for localisation.}
\label{fig:graph_propagation}
\vspace{-0.5cm}
\end{figure*}

\subsection{Graph Construction} \label{sec:graph_construction}
Our constructed graphs should represent both scene objects and their spatial relationships. Given an image $\mathbf{I}$ composed of $N$ objects, and with known intrinsic matrix $\mathbf{K}$, our Graph Constructor (shown in Fig.~\ref{fig:model_arch}) constructs a Euclidean graph $\mathcal{G} = (\mathcal{V},\mathcal{E})$, with $\mathcal{V}$ its set of $N = |\mathcal{V}|$ nodes and $\mathcal{E}$ its set of $E = |\mathcal{E}|$  edges. Each node has explicit Euclidean positional features $p_i \in \mathbb{R}^2$ representing the object's estimated location in the ground plane. Aside from its Euclidean representation, each node and edge has a set of image features $S_i^v$ and $S_i^e$, respectively.

Unlike the majority of graph-based learning methods where the input graphs are given, we construct ours on the fly. 
We design our graph based on two key choices: (1) Feature assignment: what features should each node and edge possess? (2) Graph connectivity: how should nodes be connected to each other?  


\textbf{Feature extraction:} The features assigned to a node form an initial embedding that is updated in order to make positional predictions in BEV (the same principle also applies to edge embeddings). As described in Sec.~\ref{sec:graph_motivation}, one way to determine the relative depths of objects in a scene is by comparing their appearance, geometry and position. We represent each of these aspects with the object's textural features, its 2D bounding box dimensions and its center.  Given an image $\mathbf{I}$, we first obtain a set of candidate object regions $B = \{b_1,...,b_n \}$ using a region proposal network. For each region, we obtain its center coordinates $p_i^{UV} \in \mathbb{R}^2$, its bounding box $b_i \in \mathbb{R}^4$, and an ROI pooled \cite{girshickICCV15fastrcnn} feature vector $f_i \in \mathbb{R}^{C \times h \times w}$. Additionally, for long range vertical context to aid in depth estimation, we include a set of vertical scanlines taken across the bounding box's width which are pooled horizontally to obtain a feature vector $l_i \in \mathbb{R}^{C \times H \times 1}$. This gives us a set of objects $O = \{o_1, ..., o_n\}$, where $o_i = (p_i, s_i)$ is an object with a positional coordinate $p_i$ and a set of image feature states $S_i = \{b_i, l_i, f_i\}$. 

\textbf{Graph connectivity:} A graph's structure determines how information is propagated between its nodes and edges. 
Our input objects are detected in the image, however we want to localize them in BEV. When determining the depth of an object, it is typically more helpful to determine this relative to other objects at similar depths rather than at larger depth differences. For instance, when localizing a distant pedestrian, knowing that it is behind a vehicle that is close to the camera is not useful. Instead, we estimate its depth by looking at where it is in relation to other objects at similar depths. From this, we extract two principles for our graph connectivity: (1) we want our graph structure to exist in the Euclidean domain, where we can connect objects in the image or BEV-plane, (2) we want to order objects by a coarse approximation of their relative depths and then connect them to their nearest depth neighbors. We approximate these coarse, unscaled relative depths $Z^0 = \{z_i^0, ..., z_n^0\}$ as $Z^{0} = \{\mathbf{d}_i \cdot  \mathbf{c} | i \in N\}$, where $\mathbf{c}$ is the vector from the center bottom edge of the image to its principal point $(u_0, v_0)$ and $\mathbf{d}_i$ the object's vector from  from its 2D image coordinate $p_i^{UV}$ to the principal point. These relative depth approximations are unscaled and very coarse, serving only to distinguish between an object that is obviously far and one that is near. We then use these coarse unscaled approximations $Z^0$ to generate the graph's connectivity based on nearest neighbors, resulting in an adjacency matrix $A \in \mathbb{R}^{N \times N}$ where $A_{ij}=1$ if there exists an edge between nodes $i$ and $j$. 

\textbf{Embedding initialization:} With the graph's structure determined, we assign its nodes and edges initial features which will be used for propagation later. Below, we describe the features assigned to each node followed by edges. 

First, we obtain Euclidean representations $p_i^{BEV} \in \mathbb{R}^2$ of each node. For each object, we define its initial BEV Euclidean position as $p_i^{BEV} = (z_i^0 tan^{-1}(\alpha_i^0), z_i^0)$, where $\alpha_i^0$ is its estimated viewing angle. Again, the scale of both parameters is arbitrary and only serves to indicate relative differences between positions.
We then define each node's initial feature as a tuple:
\begin{equation}\label{eq:node_features}
    \boldsymbol{v}_i^0 = (p_i, S_i = \{b_i, l_i, f_i\}),    
\end{equation}
where $p_i$ is its BEV coordinate estimate and $S_i$ a set of features from the image and BEV.

Then, for each edge's features $\boldsymbol{e}_{ij}$, we use the adjacency matrix $A$ and follow the same feature extraction process as defined above for the nodes  resulting in Eq.~\ref{eq:node_features}. Just as each object is defined by a bounding box, we similarly define each edge as its bounding box region in the image. Consequently, the node feature extraction process described above can be applied to each edge.

Finally, we have a graph $\mathcal{G} = (\mathcal{V},\mathcal{E})$ where each node has object features $\boldsymbol{v}_i^0 = (p_i^v, S_i^v = \{b_i^v, l_i^v, f_i^v\})$ and each edge has scene features $\boldsymbol{e}_{ij}^0 = (p_i^e, S_i^e = \{b_i^e, l_i^e, f_i^e\})$.

\subsection{Graph propagation} \label{sec:graph_propagation}
Given our input graph $\mathcal{G}$, with initial node features $\boldsymbol{v}_i^0$ and edge features $\boldsymbol{e}_{ij}^0$, we want to pass messages across and between both nodes and edges to learn updated embeddings $\boldsymbol{v}_i'$ and $\boldsymbol{e}_{ij}'$, which we will use for localization. We think of the input graph as a perturbed mass and spring system, and the message passing as a relaxation process that returns the system to equilibrium, which in our case would be the output object locations. Our message passing mechanism, illustrated in Fig.~\ref{fig:graph_propagation}, is motivated by the challenges posed by standard GNNs: namely, (1) the need for spatial awareness, or more explicitly, positional equivariance and (2) mutually enhanced node and edge embeddings.

\textbf{Positional equivariance:} To describe how we build positional equivariance into our graph propagation, we begin with the standard GCN formulation and the challenges it poses for our task. We then formulate our approach in response to those challenges.

Given a graph $\mathcal{G}=(\mathcal{V},\mathcal{E})$, a standard GCN layer takes as input a set of node embeddings $\{\boldsymbol{h}_i \in \mathbb{R}^d|i \in V\}$ and edges $E$. The layer produces an updated set of node embeddings $\{\boldsymbol{h}_i' \in \mathbb{R}^d|i \in V\}$ by applying the same parametric function $f_{W}$ to every node given its neighbors $\mathcal{N}_i=\{j \in V | (j,i) \in E\}$:
\begin{equation}
    \boldsymbol{h}_i' = f_{W}(\boldsymbol h_i, \textsc{aggregate}(\{\boldsymbol{h}_j |j \in \mathcal{N}_i\}))
\end{equation}
These graph convolution functions are designed to be
invariant to node position and permutation because in most graph learning tasks there is no canonical position for its nodes. Thus, such a function would fail to differentiate isomorphic nodes with the same 1-hop local neighborhood. This is unsuitable for our task, as objects may typically have the same node degree yet have completely different Euclidean structures. To overcome this, we concatenate Euclidean positional information to the node's features during message passing (Eq.~\ref{eq:gcn_w_pos}). This allows us to learn node representations which capture Euclidean structure. Additionally, the Euclidean structure of our graph changes between successive message passing layers, raising the need for updated positional information at each message passing layer. We do this by propagating positional embeddings to learn better structures at the next update (Eq. \ref{eq:mp_pos}). This differs from existing GCNs which integrate positional information by concatenating it with the input node features only \cite{dwivedi2020generalization, DBLP:conf/icml/BeainiPLHCL21, kreuzer2021rethinking}. Given these requirements, we propose a spatially-aware message-passing mechanism with learnable structural and positional embeddings. The generic update equations of our method are defined as:
\begin{equation}\label{eq:gcn_w_pos}
\begin{split}
    \boldsymbol h_i' =  f_{h}([\boldsymbol h_i \| \boldsymbol p_i], \textsc{aggregate}(\{[\boldsymbol h_j \| \boldsymbol p_j] |j \in \mathcal{N}_i\})), 
    \\
    \forall \boldsymbol h_i \in S_i.
\end{split}
\end{equation}
\begin{equation}\label{eq:mp_pos}
    \boldsymbol p_i' = f_{p}(\boldsymbol p_i, \textsc{aggregate}(\{\boldsymbol p_j |j \in \mathcal{N}_i\})),
\end{equation}
where $\|$ is the concatenation operation and $f_h$ and $f_p$ represent separate parametric functions for each feature state in $S_i$ and position, respectively. In this way, positional information is updated and dissipated through each feature state at each round of message passing.

\textbf{Mutually enhanced node and edge embeddings:} We wish to learn both node and edge embeddings by propagating information between them. That is, alongside node-to-node and edge-to-edge propagation, we also want edge-to-node and node-to-edge communication. Here we detail the application of our generic update equations (Eq.~\ref{eq:gcn_w_pos} and \ref{eq:mp_pos}) to mutually enhance node and edge embeddings. As shown in Fig.~\ref{fig:graph_propagation}, each round of message passing consists of two update mechanisms: a node-level update followed by an edge-level update. 

In our node-level update, each node state calculates a weighted average of the node and edge states of its neighborhood. Each node $i$ updates its position $\boldsymbol p_i^v$ and its feature states $\boldsymbol x_i^v \in S_i^v$ by computing a weighted average of the corresponding states of its neighborhood's nodes $\boldsymbol x_j^v \in S_j^v$ and edges $\boldsymbol x_{ij}^e \in S_{ij}^e$:
\begin{equation} \label{eq:node_state_update}
\begin{split}
        \boldsymbol x^{v \prime}_i = \alpha_{i,i} \boldsymbol \Theta_x[\boldsymbol x_i^v \| \boldsymbol p_i^v] +
        \sum_{j \in \mathcal{N}(i)} \alpha_{i,j} \boldsymbol \Theta_x([\boldsymbol x_j^v \| \boldsymbol p_j^v] + [\boldsymbol x_{ij}^e \| \boldsymbol p_{ij}^e]),
        \\ \boldsymbol x_i^v, \boldsymbol p_i^v, \boldsymbol x_{ij}^e, \boldsymbol p_{ij}^e \in \mathbb{R}^d, \boldsymbol \Theta_x \in \mathbb{R}^{d' \times 2d}
\end{split}
\end{equation}
\vspace{-0.3cm}
\begin{equation} \label{eq:node_pos_update}
\begin{split}
        \boldsymbol p^{v \prime}_i = \alpha_{i,i} \boldsymbol \Theta_p \boldsymbol p_i^v +
        \sum_{j \in \mathcal{N}(i)} \alpha_{i,j}\mathbf{\Theta_p}(\boldsymbol p_j^v + \boldsymbol p_{ij}^e),
        \\ \boldsymbol p_i^v, \boldsymbol p_{ij}^e \in \mathbb{R}^d, \boldsymbol \Theta_p \in \mathbb{R}^{d' \times d}
\end{split}
\end{equation}
where $\boldsymbol \Theta$ is a linear transformation weight matrix, and $\alpha_{i,j}$ is an attention coefficient which we define below. We have excluded the nonlinearity that is applied to $\boldsymbol x_i^{v'}$ and $\boldsymbol p_i^{v'}$ for clarity. The inclusion of  edge embedding $[\boldsymbol x_{ij}^e \| \boldsymbol p_{ij}^e]$ makes this equation a node-to-node + edge-to-node update, meaning the updated node embedding contains context from its edge embeddings. If we wanted to keep this purely node-to-node, we would simply omit the edge embeddings.

For our weighted average, we apply attention over the neighborhood similar to GAT \cite{velivckovic2018graph}, however we also include edge embeddings in this calculation. In our approach, each node $i$ calculates the importance of its neighboring node $j$ and the edge between them $e_{ij}$ using a scoring function $\Lambda: \mathbb{R}^d \times \mathbb{R}^d \times \mathbb{R}^d \rightarrow \mathbb{R}$ to calculate the attention coefficients:
\begin{equation}
\begin{split}
    \Lambda(\boldsymbol h_i,\boldsymbol h_j,\boldsymbol e_{i,j}) =  \sigma(\boldsymbol a^\top \cdot [\boldsymbol \Theta \boldsymbol h_i \| \boldsymbol \Theta \boldsymbol h_j \| \boldsymbol \Theta \boldsymbol e_{i,j}]),
    \\
    \boldsymbol \Theta \in \mathbb{R}^{d' \times d}, \boldsymbol a \in \mathbb{R}^{3d'}
\end{split}
\end{equation}
where $\boldsymbol a$ and $\boldsymbol \Theta$ are learned weights and $\sigma$ is a LeakyReLU nonlinearity. Finally, the attention coefficients are normalized across all choices of $j$ using the softmax function:
\begin{equation} \label{eq:node_attention}
    \alpha_{i,j} =
    \frac{
    \exp\left(\Lambda(\boldsymbol h_i,\boldsymbol h_j,\boldsymbol e_{i,j})\right)}
    {\sum_{k \in \mathcal{N}(i) \cup \{ i \}}
    \exp\left(\Lambda(\boldsymbol h_i,\boldsymbol h_k,\boldsymbol e_{i,k})\right)},
\end{equation}

Similarly, our edge-level update aggregates its neighboring edge and node embeddings. However, incorporating node embeddings into the aggregation function is challenging over the input graph $\mathcal{G}$. Instead, we construct its conjugate graph $\phi(\mathcal{G})$ and perform the edge-level updates here. The conjugate, $\phi(\mathcal{G})$, is a graph whose nodes are edges of $\mathcal{G}$ and two nodes are adjacent in $\phi(\mathcal{G})$ if and only if the corresponding edges are adjacent in $\mathcal{G}$. The adjacency matrix $A^e$ of the conjugate $\phi(\mathcal{G})$ can be calculated as follows:
\begin{equation}
    A^e = C^TC - 2I
\end{equation}
where $C$ is the incidence matrix of the input graph $\mathcal{G}$, and $I$ is the identity matrix. Using the conjugate $\phi(\mathcal{G}) = (\mathcal{V}_\phi, \mathcal{E}_\phi)$, each edge embedding $\{\boldsymbol e_{i,j} \in \mathbb{R}^d| i \in \mathcal{V}_\phi \}$ can be updated using Eq.~\ref{eq:node_state_update} - \ref{eq:node_attention}, except with nodes and edges swapped.

After multiple rounds of message-passing, our graph propagation method outputs a graph with updated node embeddings $\{\boldsymbol v_i{'} = (p_i^v{'}, S_i^v{'} = \{b_i^v{'}, l_i^v{'}, f_i^v{'}\}) | i \in \mathcal{V}\}$ and edge embeddings $\{\boldsymbol e_{ij}' = (p_i^e{'}, S_i^e{'} = \{b_i^e{'}, l_i^e{'}, f_i^e{'}\}) | (i,j) \in \mathcal{E}\}$. 


\subsection{Scene estimation} \label{sec:map_estimation}
Using the node embeddings output by our graph propagation module, we make predictions for scene categories. Here we take a texture-based approach based on \cite{saha2021translating}, where image features from our frontend are transformed into the BEV-plane to generate BEV maps for scene categories. However, one key difference is that we condition the latent features of this module on our node embeddings. In this way, we constrain the latent BEV space for scene categories as objects provide strong cues for the existence of roads, pavements, etc.

Following \cite{saha2021translating}, a Transformer $\mathcal{T}$ \cite{vaswani2017attention} maps image features $\mathbf{H}^{I}$ to BEV-features $\mathbf{H}^{BEV}$:
\begin{equation}
\begin{gathered}
    \mathcal{T}: \mathbb{R}^{C \times H \times W} \times \mathbb{R}^{3 \times 3} \rightarrow \mathbb{R}^{C \times 100 \times 100}
    \\
    (\mathbf{H}^{I}, \mathbf{K}) \rightarrow \mathbf{H}^{BEV}.
\end{gathered}
\end{equation}
A decoder $\mathcal{D}$ with deep layer aggregation \cite{yu2018deep} then generates BEV maps for scene categories using the latent BEV features $\mathbf{H}^{BEV}$ conditioned on node embeddings $\{\boldsymbol v_i' | i \in \mathcal{V}\}$:
\begin{equation}
\begin{gathered}
    \mathcal{D}: \mathbb{R}^{C \times 100 \times 100} \times \mathbb{R}^{L_V} \rightarrow \mathbb{R}^{k \times 100 \times 100}
    \\
    (\mathbf{H}^{BEV}, \{\boldsymbol v_i | i \in \mathcal{V}\}) \rightarrow \mathbf{M}^{BEV}.
\end{gathered}
\end{equation}

\color{black}

\subsection{Loss} \label{sec:losses}
\textbf{Object Network:} Our object network predicts parameters to help recover the object's BEV-bounding box with its yaw $\theta$, dimensions $\delta = (l, w)$, centroid $p^v = (x, z)$ and label $c$. Additionally, we also predict the midpoint of our graph's edges $p^e = (x, z)$. For each of these heads we first apply a two-layer MLP to the appropriate input feature.

\textbf{Node and Edge localization:} We supervise the graph's output node and edge embeddings $\boldsymbol v_i'$ and $\boldsymbol e_{ij}'$ entirely for localization. To obtain BEV positions $(x, z)$, we regress parameters for each using separate MLPs. We regress viewing angle $\alpha$ using the viewing angle $\alpha^0$ estimated in the Graph Constructor, and directly regress z-axis depth $z$. The viewing angle and z-axis depth are then used to determine $x$.

\textbf{Classification, Dimensions and Orientation:}
We use the initialized node features $\boldsymbol v_i^0$ to make predictions for the object's label, dimensions and orientation. As shown in Fig.~\ref{fig:model_arch}, 
these estimations are made early in the network so they constrain the graph's initial features when used for localization later. For object classification, we use the focal loss \cite{lin2017focal} in its original formulation. We directly regress object dimensions. Instead of predicting object yaw $\theta$, we predict its observation angle $\beta$, which is the sum of its viewing angle and object yaw. Estimating observation angle rather than yaw helps account for an object's changing appearance based on its viewing angle \cite{mousavian20173d}. We follow \cite{mousavian20173d} and estimate observation angle using a discrete-continuous loss: the orientation range is discretized into multiple bins and then the angle is regressed as an offset from the bin center. 

Our object network is trained using a multitask loss defined as:
\begin{equation}
    L_{total} = L_{loc^v} + L_{loc^e} + L_\theta + L_{dim} + L_c
\end{equation}
where $L_{loc^v}$, $L_{loc^e}$ and $L_{dim}$ and regression losses for the object's centroid, edge midpoint, and object dimensions, $L_\theta$ is the discrete-continuous loss for orientation, and $L_c$ is the object classification loss. All regression losses use the Smooth L1 Loss and all classification losses use a cross-entropy loss unless otherwise specified.

\textbf{Scene Network:} We train our scene network using the same multiscale Dice loss as \cite{saha2021translating}. Please see the supplementary for further details on all losses.

\begin{figure*}[t]
\centerline{\includegraphics[page=2,trim=0 2325 120 0,clip,width=0.86\linewidth]{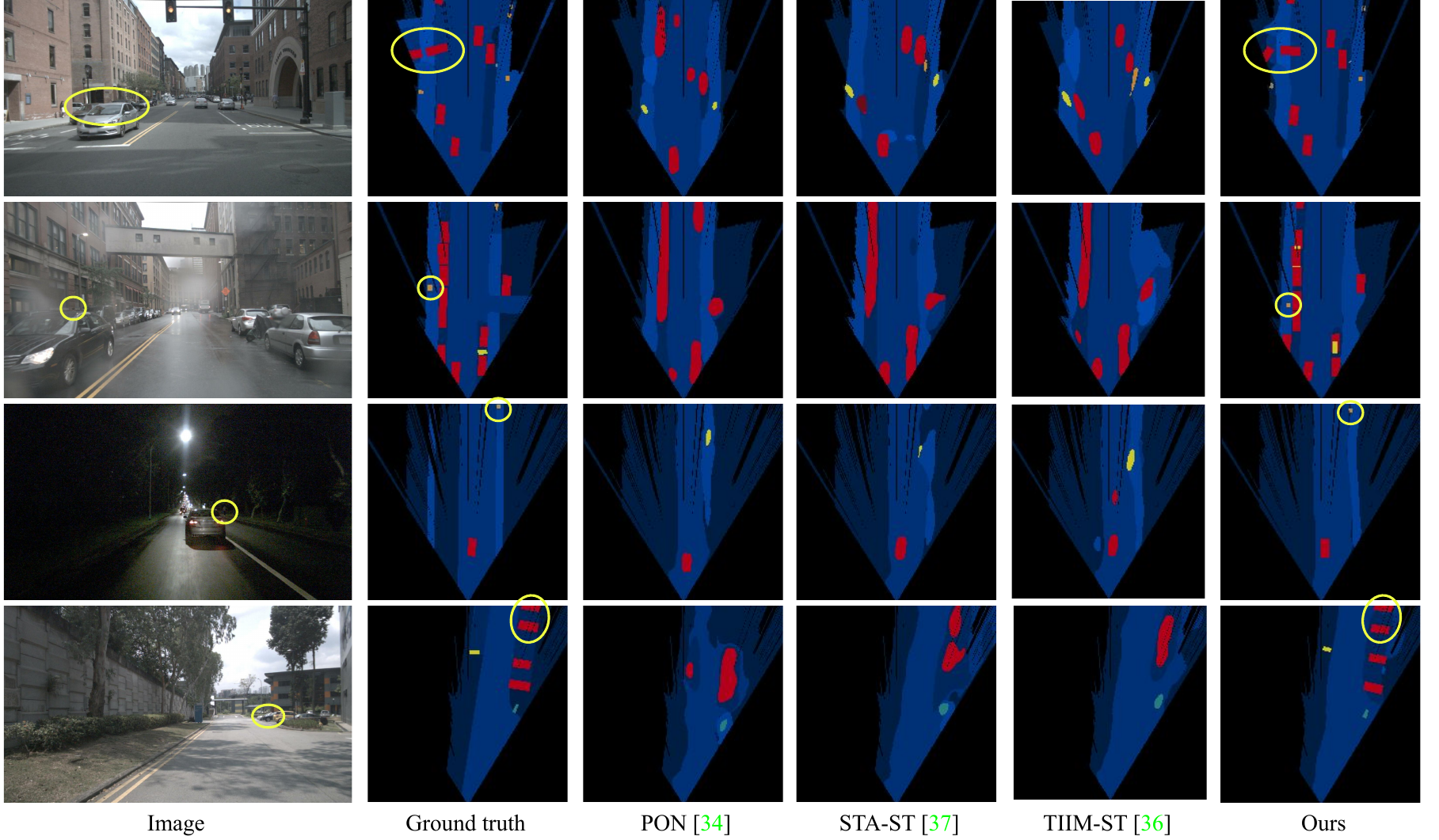}}
\vspace{-0.3cm}
\caption{Our model results on nuScenes. As highlighted, our model is able to localize distant and/or heavily occluded vehicles and pedestrians, which the other methods miss.}
\label{fig:results_nuscenes}
\vspace{-0.5cm}
\end{figure*}

\section{Experiments}
\textbf{Datasets:} We compare our approach to current state-of-the-art approaches on the nuScenes \cite{caesar2020nuscenes}, Argoverse \cite{chang2019argoverse} and Lyft \cite{kesten2019lyft} datasets. nuScenes \cite{caesar2020nuscenes} consists of 1000 clips 20-seconds in length, captured across different cities. Each scene is annotated with 3D bounding boxes for 10 object classes, along with vector maps for the road, sidewalk, and more. We generate our BEV ground truth maps following \cite{Roddick_2020_CVPR}. 

\textbf{Implementation:} We use a pretrained ResNet-50 \cite{he2016deep} with a feature pyramid (FPN) as our frontend. We extract each level of the FPN, interpolate to the same size and add, to obtain a single set of features. To obtain our candidate 2D regions in the image, we use FCOS3D \cite{wang2021fcos3d} and fine tune it to the appropriate dataset (during training we use jittered ground truth regions). Each node in our graph is connected to its 3 nearest neighbors, and we use 2 layers of message-passing. Our BEV Estimation module uses a BEV latent feature space of 100x100 pixels, with each representing $0.5m^2$ in world coordinates. Its largest scale output is 100x100 pixels which we upsample to 200x200 for fair comparison with the literature. We optimize using Adam, with a weight decay of $1e-4$, and a learning rate of $5e{-5}$ which decays by 0.99 every epoch for 50 epochs.

\subsection{Ablations}
\textbf{Mutually enhanced embeddings:} In Table \ref{tab:ablation_graph_prop} we demonstrate the effectiveness of using node and edge embeddings, mutually enhancing them, and the effect of their supervision. Beginning with a node-only graph results in the lowest IoU (although this is still higher than current SOTA BEV estimators in Table \ref{tab:results_nuscenes}). Adding edge embeddings and then allowing nodes to gather information from them (n2n + e2n) increases IoU slightly. This increase is understandable as each node is now gathering context from both its neighboring objects and its surrounding scene. The largest increase in IoU arrives from edge embeddings communicating between themselves and their supervision (n2n + e2n + e2e). This demonstrates the benefit of using edges to geometrically constrain the spatial layout of the nodes. Finally, combining all types of graph propagation with both node and edge supervision results in the best localization, highlighting the benefits of mutually enhanced embeddings.

\begin{table}
\centering
\caption{Graph propagation types. IoU(\%) across objects for different graph propagation types on nuScenes. n2n = node-to-node, e2n = edge-to-node, e2e = edge-to-edge and n2e = node-to-edge.}
\label{tab:ablation_graph_prop}
\begin{tabular}{llc}
\hline
Graph Propagation     & Supervision     & \multicolumn{1}{l}{Objects Mean} \\ \hline
n2n                   & nodes           & 20.0                            \\
n2n + e2n             & nodes           & 21.1                            \\
n2n + e2n + e2e       & nodes and edges & 25.9                            \\
n2n + e2n + e2e + n2e & nodes and edges & \textbf{27.1}                  
\end{tabular}
\vspace{-0.5cm}
\end{table}

\textbf{Graph node and edge features:} In Table \ref{tab:ablation_graph_features} we demonstrate the effect of initializing nodes and edges with different feature types. Interestingly, relying solely on appearance results in a baseline IoU of 22.5\%. This suggests the model is able to localize objects to a large extent just by comparing their textural features.
Part of this result also stems from the scene context aggregated by our frontend's feature pyramid, meaning the ROI crops of each object contain broader scene context and not just the object's appearance. This may explain why the inclusion of scanline features do not improve upon this much. In contrast, including the object's bounding box parameters creates the largest improvement. Since images of urban driving environments display strong regularities in scene structure, 
knowing where an object is in the image and its image dimensions provides enough information to infer depth coarsely. Finally, modulating each feature type with positional information improves upon this further. This can be explained by the attention mechanism: 
positional approximations of a node's neighborhood can signal which may be the most relevant, as it is typically the nearest neighbors which are of most use in terms of context for localisation.

\begin{table}
\centering
\caption{IoU (\%) effect of graph feature types on nuScenes.}
\label{tab:ablation_graph_features}
\begin{tabular}{lc}
\hline
Graph Node and Edge Features               & \multicolumn{1}{l}{Objects Mean} \\ \hline
Appearance                                 & 22.5                            \\
Appearance, scanline                        & 22.9                            \\
Appearance, geometry                       & 26.1                            \\
Appearance, geometry, scanline             & 26.2                            \\
Position w. appearance, scanline, geometry & \textbf{27.1}              
\end{tabular}
\vspace{-0.5cm}
\end{table}

\color{black}
\begin{table*}[t]
\centering
\caption{nuScenes IoU (\%) results on the validation split of \cite{Roddick_2020_CVPR}. The last row displays the relative improvement per category over the current state-of-the-art BEV Estimators.}
\label{tab:results_nuscenes}
\resizebox{1.04\textwidth}{!}{%
\begin{tabular}{l|cccccccccccccc|cc}
Model   & \multicolumn{1}{l}{Drivable} & \multicolumn{1}{l}{Crossing} & \multicolumn{1}{l}{Walkway} & \multicolumn{1}{l}{Carpark} & \multicolumn{1}{l}{Car} & \multicolumn{1}{l}{Truck} & \multicolumn{1}{l}{Trailer} & \multicolumn{1}{l}{Bus} & \multicolumn{1}{l}{Con.Veh.} & \multicolumn{1}{l}{Bike} & \multicolumn{1}{l}{Motorbike} & \multicolumn{1}{l}{Ped.} & \multicolumn{1}{l}{Cone} & \multicolumn{1}{l|}{Barrier} & \multicolumn{1}{l}{Mean} & \multicolumn{1}{l}{Objects Mean} \\ \hline
VPN \cite{pan2020cross}    & 58.0                         & 27.3                         & 29.4                        & 12.3                        & 25.5                    & 17.3                      & 16.6                        & 20.0                    & 4.9                          & 4.4                         & 5.6                            & 7.1                      & 4.6                      & 10.8                         & 17.4                     & 11.7                            \\
PON \cite{Roddick_2020_CVPR} & 60.4                         & 28.0                         & 31.0                        & 18.4                        & 24.7                    & 16.3                      & 16.6                        & 20.8                    & 12.3                         & 9.4                         & 7.0                            & 8.2                      & 5.7                      & 8.1                          & 19.1                     & 12.9                            \\
STA-S \cite{saha2021enabling}  & 71.1                         & 31.5                         & 32.0                        & 28.0                        & 34.6                    & 18.0                      & 11.4                        & 22.8                    & 10.0                         & 14.6                        & 7.1                            & 7.4                      & 5.8                      & 10.8                         & 21.8                     & 14.3                            \\
TIIM-S \cite{saha2021translating}  & 72.6                         & 36.3                         & 32.4                        & 30.5                        & 37.4                    & 24.5                      & 15.5                        & 32.5                    & 14.8                         & \textbf{15.1 }                       & 8.1                            & 8.7                      & 7.4                      & 15.1                         & 25.1                     & 17.9                            \\
FCOS3D \cite{wang2021fcos3d}  & -                          & -                          & -                         & -                         & 28.6                    & 25.0                      & 20.4                        & 34.2                    & 8.1                          & 11.1                        & 14.6                           & 9.8                      & \textbf{9.5}                      & 23.9                         & -                    & 18.6                            \\
STA-ST \cite{saha2021enabling}  & 70.7                         & 31.1                         & 32.4                        & 33.5                        & 36.0                    & 22.8                      & 13.6                        & 29.2                    & 12.1                         & 12.1                        & 8.0                            & 8.6                      & 6.9                      & 14.2                         & 23.7                     & 16.4                            \\
TIIM-ST \cite{saha2021translating}  & 74.5                         & 36.6                         & 35.9                        & 31.3                        & 39.7                    & 26.3                      & 13.9                        & 32.8                    & 14.2                         & 14.7                        & 7.6                            & 9.5                      & 7.6                      & 14.7                         & 25.7                     & 18.1                            \\ \hline
Ours      & \textbf{75.9}                & \textbf{39.7}                & \textbf{37.9}               & \textbf{36.8}               & \textbf{41.5}           & \textbf{38.0}             & \textbf{28.8}               & \textbf{58.1}           & \textbf{23.8}                & 12.2                        & \textbf{18.4}                  & \textbf{11.5}            & 9.0             & \textbf{30.1}                & \textbf{33.0}            & \textbf{27.1} \\          
 Rel. improv. (\%)  & 6.4                         & 12.9                        & 21.6   & 20.0	& 4.5	& 44.5	& 73.2	& 77.2	& 60.7	& -19.0	& 127.3	& 21.5	& 17.8	& 99.7	& 32.4	& 50.0     \\
\end{tabular}%
}
\vspace{-0.5cm}
\end{table*}

\begin{table}
\centering
\caption{Argoverse results on the validation split of \cite{Roddick_2020_CVPR}.}
\label{tab:results_argoverse}
\resizebox{0.5\textwidth}{!}{%
\begin{tabular}{l|cccccccc|c}
Method  & \multicolumn{1}{l}{Drivable} & \multicolumn{1}{l}{Veh.} & \multicolumn{1}{l}{Ped.} & \multicolumn{1}{l}{L.Veh.} & \multicolumn{1}{l}{Bike} & \multicolumn{1}{l}{Bus} & \multicolumn{1}{l}{Trailer} & \multicolumn{1}{l|}{Motorbike} & \multicolumn{1}{l}{Mean} \\ \hline
VPN \cite{pan2020cross}     & 64.9                         & 23.9                        & 6.2                            & 9.7                                & 0.9                         & 3.0                     & 0.4                         & 1.9                             & 13.9                     \\
PON \cite{Roddick_2020_CVPR} & 65.4                         & 31.4                        & \textbf{7.4}                            & 11.1                               & 3.6                         & 11.0                    & 0.7                         & 5.7                             & 17.0                     \\
TIIM-S \cite{saha2021translating}  & 75.9                         & 35.8                        & 5.7                            & 14.9                               & \textbf{3.7}                         & 30.2                    & 12.1                        & 2.6                             & 22.6                     \\
Ours    & \textbf{78.2}                & \textbf{52.1}               & 6.9                   & \textbf{23.0}                      & 3.1                & \textbf{49.0}           & \textbf{23.8}               & \textbf{6.9}                    & \textbf{30.3}           
\end{tabular}
}
\end{table}

\begin{table}
\centering
\caption{IoU(\%) against `adaptive' approaches on nuScenes canonical validation split and Lyft.}
\label{tab:us_vs_philion}
\resizebox{0.45\textwidth}{!}{%
\begin{tabular}{l|lll|lll}
        & \multicolumn{3}{c|}{nuScenes}                                                                     & \multicolumn{3}{c}{Lyft}                                    \\ \cline{2-7} 
        & \multicolumn{1}{c}{Driv.} & \multicolumn{1}{c}{Car} & \multicolumn{1}{c|}{Veh.} &
        \multicolumn{1}{c}{Driv.} & \multicolumn{1}{c}{Car} & \multicolumn{1}{c}{Veh.} \\ \hline
LSS \cite{philion2020lift} & 72.9  & 32.0 & 32.0 & -  & 43.1 & 44.6                                 \\
FIERY \cite{fiery2021} & -  & 39.9 & - & -  & - & -                                 \\
TIIM-S \cite{saha2021translating}      & 78.9  & 39.9  & 38.9 &  82.0 &   45.9  & 45.4 \\ 
Ours & \textbf{81.4} & \textbf{41.7} & \textbf{49.8} & \textbf{84.2} & \textbf{47.4} & \textbf{48.3}\\
\hline
\end{tabular}
}
\vspace{-0.3cm}
\end{table}

\textbf{Effect of graph node degree:} In Table \ref{tab:ablation_node_degree} we examine the effect of node degree when constructing our input graphs. Broadly, IoU is inversely proportional to node degree. 
The progressive decrease in performance is explained by the information available to each node when aggregating its neighborhood: larger node degrees entail more redundancies, and learning to minimize this is increasingly challenging as redundancies in the neighborhood grow.


\begin{table}
\centering
\caption{IoU(\%) effect of node degree on nuScenes.}
\label{tab:ablation_node_degree}
\resizebox{0.48\textwidth}{!}{%
\begin{tabular}{c|cccccccc}
Node Degree  & 0    & 1    & 2    & 3             & 5    & 10   & 15    & 20+   \\ \hline
Objects Mean & 20.2 & 26.2 & 26.5 & \textbf{27.1} & 21.1 & 14.2 & 11.56 & 10.3
\end{tabular}%
}
\vspace{-0.5cm}
\end{table}

\subsection{Comparison to SOTA}
\textbf{Baselines:} We compare against a number of state-of-the-art BEV estimation methods across nuScenes, Argoverse and Lyft datasets. We compare against `fixed' approaches VPN \cite{pan2020cross}, PON \cite{Roddick_2020_CVPR}, STA \cite{saha2021enabling}, and `adaptive' approaches LSS \cite{philion2020lift}, FIERY \cite{fiery2021} and TIIM \cite{saha2021translating}. For completeness, we also compare our BEV estimation results on `objects' to those of a state-of-the-art 3D object detector FCOS3D \cite{wang2021fcos3d}.

In Table \ref{tab:results_nuscenes}, we demonstrate a 30\% relative improvement over the next best performing method TIIM \cite{saha2021translating}, outperforming both its spatial TIIM-S and spatio-temporal TIIM-ST models. In particular, our `objects' classes show a relative increase of 50\%, with Motorbikes and Barriers showing a 100\% relative gain. Much of this difference can be attributed to our object-based approach to localizing these classes. For fair comparison, we also compare our results on `objects' to another object-based approach: FCOS3D \cite{wang2021fcos3d}. Here we demonstrate a similar 45\% relative improvement across all object classes. With FCOS3D producing object bounding boxes just like we do, the difference in performance here is likely due to our graph-based approach to localization. Our results on Argoverse display similar patterns, where we improve upon the next best performing method TIIM-S \cite{saha2021translating} by 33\%.

In Table \ref{tab:us_vs_philion} we outperform adaptive methods \cite{philion2020lift, fiery2021, saha2021enabling} on nuScenes and Lyft. A true comparison on Lyft with LSS \cite{philion2020lift} is not possible as we were unable to acquire their train/validation split.
However, the difference in performance can be attributed to us comparing objects for context, while adaptive approaches rely on scene context.

\begin{figure}
\centerline{\includegraphics[trim=0 20 0 20,width=0.6\linewidth]{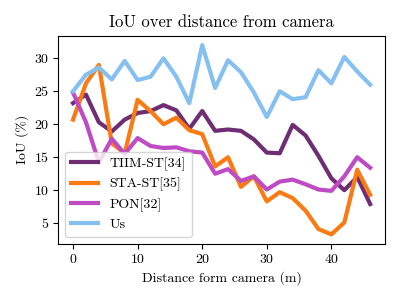}}
\vspace{-0.2cm}
\caption{IoU (\%) over distance from camera on nuScenes.}
\label{fig:results_bev_error}
\vspace{-0.5cm}
\end{figure}

To obtain a more granular understanding of our method's performance compared to SOTA, we compare IoU (\%) accuracy of `objects' as a function of distance from camera in Fig.~\ref{fig:results_bev_error}. Current SOTA BEV-Estimators generally drop in IoU as distance increases. While our method shows a slight drop in IoU between 25-45m, it is broadly maintained along the depth axis. This capability for localizing distant objects can be seen in our qualitative results in Fig.~\ref{fig:results_nuscenes}, where our model is able to correctly localize objects that are distant and/or heavily occluded, which other methods miss.

\vspace{-0.1cm}
\subsection{Limitations}
Our Graph Constructor enforces many inductive biases, in terms of graph connectivity and feature type 
. Ideally we want to jointly learn both the connectivity and feature selection while optimizing for object localization. For instance, our IoU on Bikes in Table \ref{tab:results_nuscenes} suggests our graph construction method is not optimal across all object categories. While Bikes are a difficult category due to their large variations in pose, it nonetheless presents an opportunity to learn the graph construction method.

\vspace{-0.2cm}
\section{Conclusion}
We proposed a graph convolution network with a novel position-equivariant message passing mechanism to localize objects in BEV from an image. In particular, we demonstrated the benefit of learning both node and edge embeddings and methods for their mutual enhancement. One of our key insights for better localization is the use of edge features as a method of gathering scene context and its supervision as a way of placing geometric constraints on object locations. Our models are state-of-the-art in BEV estimation from monocular images across three large-scale datasets.

\section*{Acknowledgements}
This project was supported by the EPSRC project ROSSINI (EP/S016317/1) and studentship 2327211 (EP/T517616/1).

{\small
\bibliographystyle{ieee_fullname}
\bibliography{ms.bib}
}
\end{document}


\appendix
\section{Losses}
\textbf{Object Network:} Our object network consists of losses applied at different layers within it, as shown in Fig.~1 of the main paper. The initial node embeddings $\{\boldsymbol v_i^0 | i \in \mathcal{V}\}$ of input graph $\mathcal{G}$ are supervised for object yaw $\theta$, dimensions $\delta = (l, w)$ and label $c$. After message-passing across this graph, the updated node embeddings $\{\boldsymbol v_i' | i \in \mathcal{V}\}$ are used for the object's centroid $p^v = (x, z)$, while the updated edge embeddings $\{\boldsymbol e_{ij}' | (i,i) \in \mathcal{E}\}$ are trained to predict the midpoint of the graph's edges, that is, the midpoint between node $i$ and $j$: $p^e = (x, z)$. 

The object's yaw $\theta$ is a single scalar. However, as mentioned in the main paper, it is difficult to regress. Instead, we follow Mousavian \emph{et al.}\cite{mousavian20173d,zhou2019objects,liu2019deep} and predict the object's observation angle $\beta$ as a vector trained with a discrete-continuous loss. First, the observation angle $\beta$ is defined as follows:
\begin{equation}
    \beta = \alpha + \theta,
\end{equation}
where $\alpha$ is the viewing angle (the polar angle of the ray). To construct the multi-scalar $\beta$ encoding, the orientation range $[-\pi, \pi]$ is discretised into $n$ overlapping bins. Within each bin, the network estimates the confidence probability $c_i$ of the observation angle falling within the bin and the residual rotation to the bin center $m_i$. The residual rotation is represented by the sine and cosine of the offset to the bin center. This results in 3 parameters for each bin $i$: $(c_i, \mathrm{sin}(\beta_i - m_i), \mathrm{cos}(\beta_i - m_i))$. The confidence probabilities are trained with a cross-entropy loss and the residuals with a Smooth L1 Loss:
\begin{equation}
    L_\beta = \frac{1}{|\mathcal{V}|}\sum_{k=1}^{|\mathcal{V}|} \sum_{i=1}^{n} CE(\hat{c}_i,c_i) + c_i * SmoothL1(\hat{a}_i, a_i),
\end{equation}
where $CE$ is the cross-entropy loss, $c_i$ is the ground truth binary variable of the angle $\beta_i$ falling within the bin $i$, and $a_i = (\mathrm{sin}(\beta_i - m_i), \mathrm{cos}(\beta_i - m_i))$. In practice, we find $n=2$ bins sufficient.

The object's BEV dimensions $\delta = (l, w)$ are regressed directly with a Smooth L1 Loss:
\begin{equation} \label{eq:smoothl1}
    L_{dim} = \frac{1}{|\mathcal{V}|} \sum_{k=1}^{|\mathcal{V}|}
    SmoothL1(\hat{\delta}_k, \delta_k),
\end{equation}
where $\delta_k$ is the object's length and width in meters.
The object's label $c$ is supervised with a focal loss \cite{lin2017focal} for classification:
\begin{equation}
    L_c = \frac{1}{|\mathcal{V}|} \sum_{k=1}^{|\mathcal{V}|} - \alpha(1 - p_k)^\gamma \mathrm{log}(p_k),
\end{equation}
where $p_k$ is the class probability of a predicted object. We follow the hyperparameter settings of the paper, with $\alpha=0.25$ and $\gamma=2$.

To predict each object's centroid $p^v = (x, z)$, we regress its viewing angle $\alpha$ and depth $z$ directly using Eq.~\ref{eq:smoothl1}. The $x$-value of the centroid is recovered using both $\alpha$ and $z$. We follow the same procedure for the midpoint of each edge $p^e = (x, z)$. 

\textbf{Scene Network:} To supervise our scene network, we follow Saha \emph{et al.} \cite{saha2021enabling} and apply a Dice loss to each BEV map $\mathbf{M}^{BEV}_u$ generated at scale $u$. In total, the multi-scale Dice loss across all scales $U$ is defined as:
\begin{equation}
    L_{scene} = 1 - \frac{1}{C} \sum_{u=1}^U \sum_{c=1}^C \frac{2 \sum_i^N \hat{m}_i^c m_i^c}{\sum_i^N \hat{m}_i^c + m_i^c + \epsilon},
\end{equation}
where $\hat{m}_i^c$ is the predicted sigmoid output of the network at scale $u$, $m_i^c$ is the ground truth binary variable at scale $u$ and $\epsilon$ is a constant which prevents division by zero.

{\small
\bibliographystyle{ieee_fullname}
\bibliography{supplement.bib}
}